\documentclass[letterpaper]{article} 
\usepackage{aaai25}  
\usepackage{times}  
\usepackage{helvet}  
\usepackage{courier}  
\usepackage[hyphens]{url}  
\usepackage{graphicx} 
\usepackage{natbib}  
\usepackage{caption} 
\usepackage{algorithm}
\usepackage{algorithmic}
\usepackage{booktabs}
\usepackage{multirow} 
\usepackage{color}
\usepackage{amsmath}
\usepackage{amssymb}
\usepackage{pifont}
\usepackage{colortbl}
\usepackage{xcolor}
\definecolor{lightblue}{rgb}{0.93,0.95,1.0}
\usepackage{arydshln}
\usepackage{makecell}
\usepackage{url}

\usepackage{newfloat}
\usepackage{listings}
\DeclareCaptionStyle{ruled}{labelfont=normalfont,labelsep=colon,strut=off} 
\floatstyle{ruled}
\newfloat{listing}{tb}{lst}{}
\floatname{listing}{Listing}
\title{Focus on Local: Finding Reliable Discriminative Regions for \\ Visual Place Recognition}
\author{
    Changwei Wang\textsuperscript{\rm 1,\rm 3},
    Shunpeng Chen\textsuperscript{\rm 2},
    Yukun Song\textsuperscript{\rm 2},
    Rongtao Xu\textsuperscript{\rm 4},
    Zherui Zhang\textsuperscript{\rm 2},
    Jiguang Zhang\textsuperscript{\rm 4},
    Haoran Yang\textsuperscript{\rm 5},
    Yu Zhang\textsuperscript{\rm 5},
    Kexue Fu\textsuperscript{\rm 1,\rm 3},
    Shide Du\textsuperscript{\rm 6},
    Zhiwei Xu\textsuperscript{\rm 7},
    \\Longxiang Gao\textsuperscript{\rm 1,\rm 3,}\thanks{
Longxiang Gao is the corresponding author. },
    Li Guo\textsuperscript{\rm 2},
    Shibiao Xu\textsuperscript{\rm 2}
}
\affiliations{
    \textsuperscript{\rm 1}Key Laboratory of Computing Power Network and Information Security, Ministry of Education, Shandong Computer Science Center, Qilu University of Technology (Shandong Academy of Sciences)\\
     \textsuperscript{\rm 2}School of Artificial Intelligence, Beijing University of Posts and Telecommunications\\
     \textsuperscript{\rm 3}Shandong Provincial Key Laboratory of Computing Power Internet and Service Computing, Shandong Fundamental Research Center for Computer Science\\
      \textsuperscript{\rm 4}MAIS, Institute of Automation, Chinese Academy of Sciences\\
    \textsuperscript{\rm 5}Tongji University\\
     \textsuperscript{\rm 6} College of Computer and Data Science, Fuzhou University\\
    \textsuperscript{\rm 7}Shandong University\\
     gaolx@sdas.org

}

\usepackage{bibentry}

\begin{document}

\maketitle

\begin{abstract}
Visual Place Recognition (VPR) is aimed at predicting the location of a query image by referencing a database of geotagged images.
For VPR task, often fewer discriminative local regions in an image produce important effects while mundane background regions do not contribute or even cause perceptual aliasing because of easy overlap.
However, existing methods lack precisely modeling and full exploitation of these discriminative regions. 
In this paper, we propose the Focus on Local (FoL) approach to stimulate the performance of image retrieval and re-ranking in VPR simultaneously by mining and exploiting reliable discriminative local regions in images and introducing pseudo-correlation supervision.
First, we design two losses, Extraction-Aggregation Spatial Alignment Loss (SAL)
and Foreground-Background Contrast Enhancement Loss (CEL), to explicitly model reliable discriminative local regions and use them to guide the generation of global representations and efficient re-ranking.
Second, we introduce a weakly-supervised local feature training strategy based on pseudo-correspondences obtained from aggregating global features to alleviate the lack of local correspondences ground truth for the VPR task.
Third, we suggest an efficient re-ranking pipeline that is efficiently and precisely based on discriminative region guidance.
Finally, experimental results show that our FoL achieves the state-of-the-art on multiple VPR benchmarks in both image retrieval and re-ranking stages and also significantly outperforms existing two-stage VPR methods in terms of computational efficiency.
\end{abstract}


\begin{links}
     \link{Code}{https://github.com/chenshunpeng/FoL}
 \end{links}

\section{Introduction}

\begin{figure}[t!]
    \centering
    \includegraphics[width=1\linewidth]{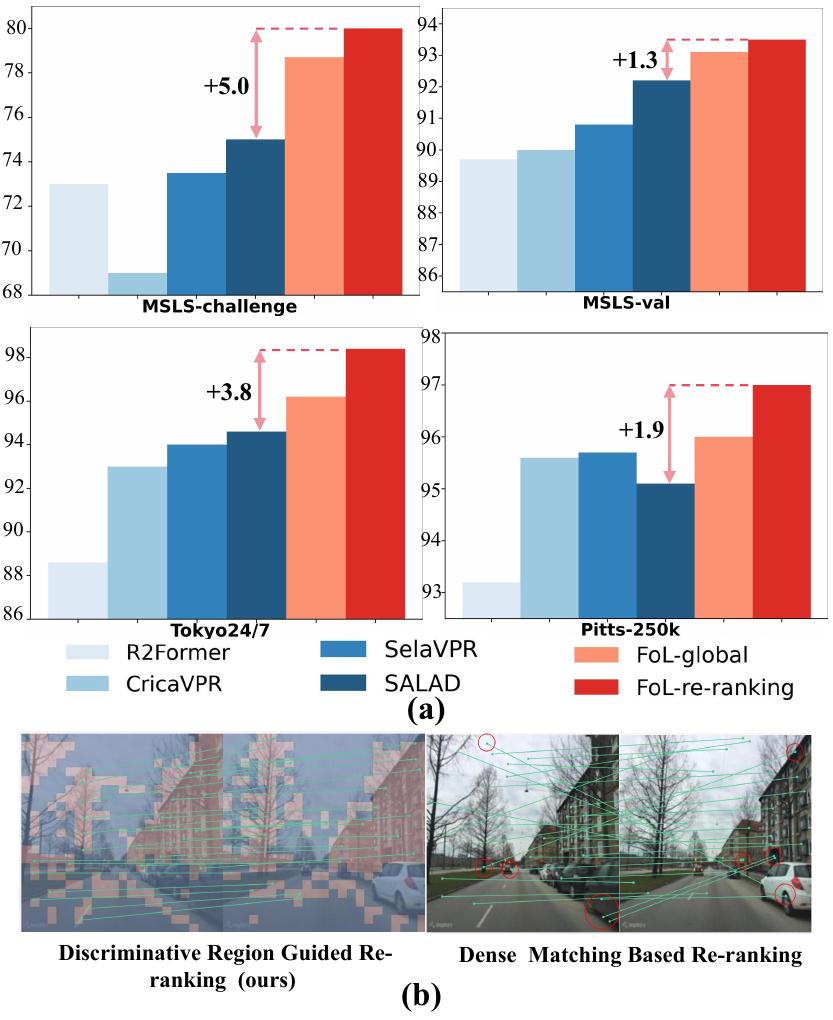}
    \caption{ (a) The results show that our FoL achieves state-of-the-art performance in the image retrieval phase alone and significantly outperforms recent methods after re-ranking.
(b) Our FoL proposes to do local matching only in the discriminative region, which not only improves the accuracy of re-ranking but also greatly improves the efficiency.
}
\label{fig:introduction}
\end{figure}

Visual Place Recognition (VPR), also known as visual geolocalization~\cite{berton2022rethinking}, is used to retrieve images of the most similar locations from geotagged databases to estimate the rough geographic location of a query image and is an essential part of many robotics~\cite{chen2017only} and computer vision~\cite{middelberg2014scalable,guo1,guo2,chang1,chang2} tasks.
Early approaches viewed VPR as a single-stage image retrieval problem~\cite{datta2008image}, using global features (\textit{a.k.a.} global descriptors) to represent the location image, similarity search in that feature space, and determining the localized location by nearest neighbor matching.
However, these global features ignore spatial information, making single-stage VPR methods based on these features prone to perceptual aliasing~\cite{salad}, \textit{i.e.}, it is difficult to distinguish highly similar images taken from different locations.

To utilize the spatial information, some recent VPR methods have proposed two-stage promising solutions~\cite{patchvlad,R2FormerCVPR2023,lu2024towards}, in which they retrieve the top \textit{k} candidates in the database using global features, and then rerank these candidates by matching local features. 
Although the two-stage methods have yielded encouraging results in terms of accuracy through the introduction of spatial information checking, there are still three problems with these methods that have not been well addressed:
\textbf{\textit{i)}} They lacked modeling and utilization of discriminatory regions, which are critical for VPR task;
\textbf{\textit{ii)}} The lack of local correspondence labels in the existing VPR dataset leads to ineffective local matching supervision of the re-ranking stage;
\textbf{\textit{iii)}} Matching on dense local features in the re-ranking stage leads to huge runtime overhead and memory usage.

To alleviate the above problems, this work advocates ``focus on local'' and proposes FoL, an efficient two-stage VPR method that makes better use of spatial local information.
First, we design two losses to explicitly model regions of the VPR task that are discriminative and reliable. 
Specifically, we propose Extraction-Aggregation Spatial Alignment Loss (SAL) to achieve mutually reinforcing self-supervised learning by aligning the regions of interest of the feature extractor (\textit{i.e.}, the self-attention map in the transformer) with the regions that are not discarded by the feature aggregator.
Besides, we introduce the Foreground-Background Contrast Enhancement Loss (CEL) to encourage  the network to generate regions that are both reliable, meaning they yield consistent attention across images, and discriminative, ensuring they are well differentiated from the background.
As shown in Figure~\ref{fig:introduction} (a), under the influence of discriminative region modeling, our FoL generates better global features, thus obtaining higher retrieval accuracy.
Second, we introduce a weakly supervised local feature training strategy to improve the performance of local matching~\cite{local1,local2} in the re-ranking stage. This strategy is based on pseudo-correspondences derived from aggregated global features, which helps to solve the problem of scarcity of ground truth for local correspondences~\cite{local3,local4}.
Third, we suggest an efficient re-ranking pipeline that is efficiently and precisely based on discriminative region guidance, as shown in Figure~\ref{fig:introduction} (b).

To summarize, our work brings the following contributions:
\textbf{1)} We design two well-designed losses for explicitly modeling reliable discriminative regions and enabling both image retrieval and local matching stages to benefit from them.
\textbf{2)} We introduce a weakly supervised local feature training strategy based on pseudo-correspondence to improve re-ranking performance.
\textbf{3)} We suggest using the discriminative region for guiding the local features matching in the re-ranking stage, thereby furthering the accuracy and efficiency.
\textbf{4)} Extensive experiments showing our FoL has a state-of-the-art performance on a wide range of VPR benchmarks with a competitive inference speed.

\section{Related Works}
\begin{figure*}[htp]
    \centering
    \includegraphics[width=1\linewidth]{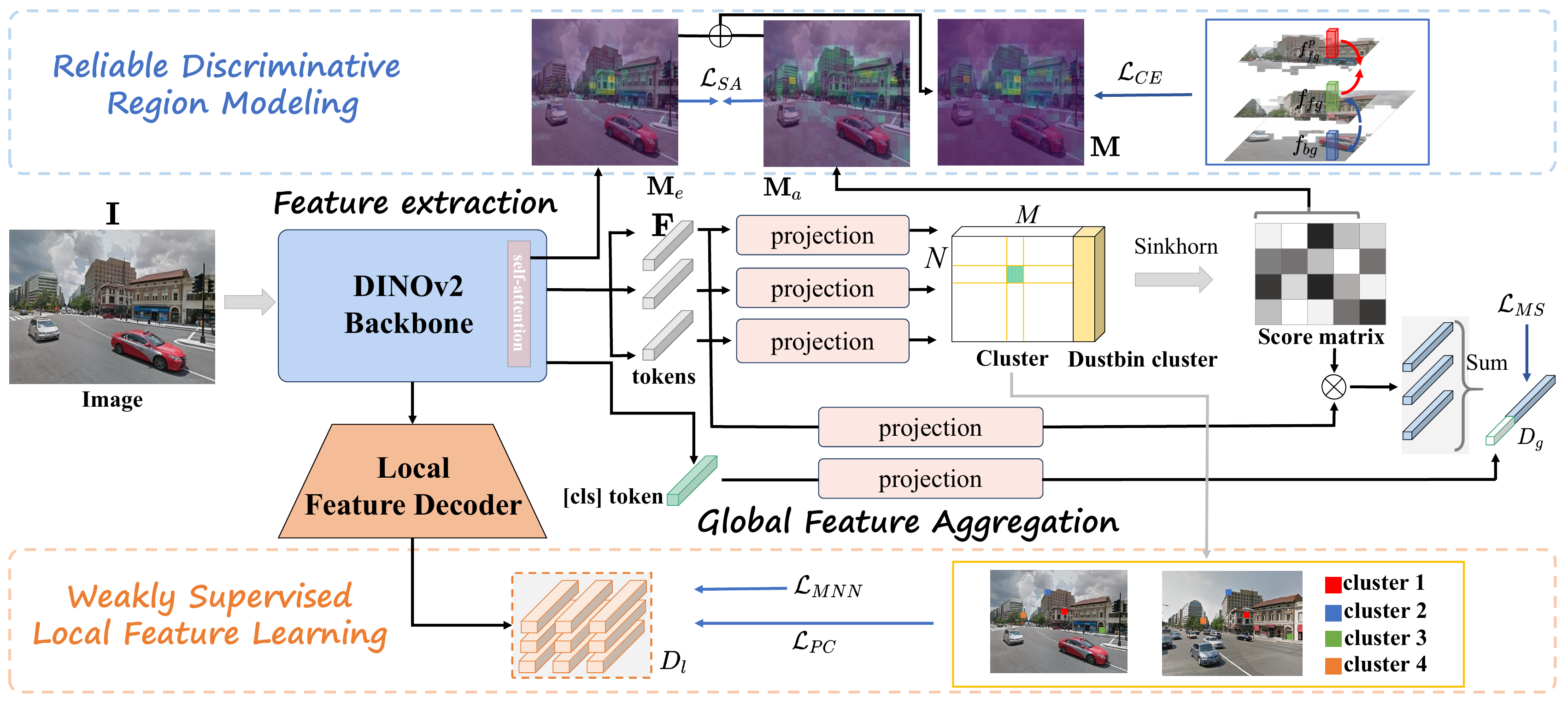}
    \caption{Illustration of our FoL's pipeline during training. 
    In addition to including common feature extraction and global feature aggregation steps, our FoL also includes Reliable discriminative region modeling and weakly supervised
local feature learning to fully introduce spatially localized information to simultaneously improve the performance of the two-stage VPR method for image retrieval and re-ranking.
}
    \label{fig:method}
\end{figure*}

\textbf{One-Stage VPR:} 
The one-stage VPR approaches produce global representations for image matching by aggregating early handcrafted local features~\cite{bay2006surf} or deep learning-based features from convolutional neural networks~\cite{netvlad}, MLPs~\cite{mixvpr}, and transformers~\cite{keetha2023anyloc}.
With the recent rise of visual foundation models (VFMs), several approaches have adopted the foundation model DINOv2~\cite{oquab2023dinov2} as the backbone of the VPR task and have shown promising performance.
One representative work is CricaVPR~\cite{crica}, which integrates adapters into DINOv2, employing a self-attention mechanism to associate multiple images within a batch, leveraging variations across images as cues to guide representation learning.
Meanwhile, SALAD~\cite{salad} redefines NetVLAD~\cite{netvlad}'s soft-assignment as the optimal transmission problem aggregating local features from DINOv2.
However, one-stage methods are prone to perceptual aliasing~\cite{sela} due to the fact that they only utilize global information for the purpose and ignore spatial information that can be used for position calibration.
In contrast, our proposed FoL models and exploits image spatial discriminative regions so that the network pays more attention to valuable regions to alleviate the perceptual aliasing of global retrieval and imposes spatial information validation through local matching between discriminative regions to further refine the global retrieval results efficiently.

\textbf{Two-Stage VPR:} 
A recent trend in VPR is the use of a two-stage retrieval strategy~\cite{patchvlad,lu2024deep}, which first performs an initial ranking based on the global representation similarity, and then re-ranks the top K retrieved candidates in the second stage using local features containing spatial information.
The two-stage VPR methods achieve a more competitive performance due to the use of spatial information.
Patch-NetVLAD~\cite{patchvlad} is a two-stage method that first retrieves candidate images using global NetVLAD features and subsequently reorders these candidates by leveraging multi-scale patch-level features derived from the NetVLAD framework.
TransVPR~\cite{transvpr}  leverages the transformer architecture's multi-level attention mechanism to derive global features for candidate retrieval, and uses an attention mask to filter the feature map to generate key patch descriptors for reordering candidates. 
R2Former\cite{R2FormerCVPR2023} proposes a unified framework for place recognition that integrates retrieval and re-ranking, where the re-ranking module considers feature correlation, attention values, and coordinates.
Recently, SelaVPR\cite{lu2024towards} integrated trainable adapters into the DINOv2 architecture then used GeM pooling techniques for feature aggregation and performed re-ranking by introducing an adapter that produces local features.
Different from existing two-stage methods, our proposed FoL explicitly models reliable discriminative regions by leveraging extraction-aggregation spatial alignment and foreground-background contrast enhancement losses. These mined regions can be used to facilitate efficient and accurate local feature matching in the re-ranking stage. Additionally, our FoL also introduce pseudo-correspondence-based weakly-supervised local feature learning techniques to enhance the accuracy of the re-ranking stage.

\section{Methodology}

\subsection{Network Architecture}
As shown in Figure~\ref{fig:method}, the image is first fed into the backbone network to extract the features, and then goes through two branches to derive the global feature and local features respectively. Specifically, global feature for image retrieval are obtained after feature aggregation, while local features are obtained by a lightweight resolution recovery decoder.

\noindent\textbf{Feature Extraction.}
Following recent works~\cite{keetha2023anyloc}, we use the transformer-based vision foundation model DINOv2~\cite{oquab2023dinov2} as a feature extractor, for input image  $\mathbf{I} \in \mathbb{R}^{h \times w \times 3}$ after feature extraction to obtain  feature $\mathbf{F}$ containing $N=h\times w/14^{2}$ tokens $\mathbf{t}_{i}, i \in \mathbb{R}^{N}$.

\noindent\textbf{Global Feature Aggregation.}
We adopt the optimal assignment based on Sinkhorn algorithm~\cite{cuturi2013sinkhorn} proposed by SALAD~\cite{salad} to aggregate the features $\mathbf{F}$ to construct the global feature $\mathbf{D}_{g}$.
Specifically, tokens from $\mathbf{F}$ are assigned to $M$ clusters according to the score matrix computed by the Sinkhorn algorithm, and an additional dustbin cluster can is set up to discard useless background region tokens. 
In particular, we refer to the tokens regions assigned to the $M$ clusters outside the dustbin cluster can as the feature aggregation attention mask $\mathbf{M}_{a}$.

\noindent\textbf{Local Feature Decoder.}
The decoder is lightweight consisting of only two Deconvolution layers and a ReLU layer.

\subsection{Reliable Discriminative Region Modeling}
\label{sec:rdr}
In FoL, we design two losses, Extraction-Aggregation Spatial Alignment Loss (SAL) and Foreground-Background Contrast Enhancement Loss (CEL), to model reliable discriminative regions.
We seek to construct discriminative regions by utilizing the spatial information capturing capabilities readily available in feature extraction and global feature aggregation of VPR pipeline.

For feature extraction, the self-attention score computation of the transformer architecture incorporates the level of attention paid to each token representing an image patch, which can be viewed as modeling spatial information. Specifically, $\mathbf{M}_{e}$ is obtained by averaging the attention scores of the last multi-head attention layer [cls] token over patch tokens.
For feature aggregation, the dustbin cluster is dedicated to hold background features, while other foreground features can be obtained by assigning to the set of $M$ clusters, thus we can obtain discriminative foreground region mask $\mathbf{M}_{a}$.
Finally, we obtain the discriminative region mask $\mathbf{M}$ by fusing $\mathbf{M}_{e}$ and $\mathbf{M}_{a}$, \textit{i.e.}, $\mathbf{M}=(\mathbf{M}_{e}+\mathbf{M}_{a})/2$.

\noindent\textbf{Extraction-Aggregation Spatial Alignment Loss.}
We propose Extraction-Aggregation Spatial Alignment Loss (SAL) to achieve mutually reinforcing self-supervised learning by aligning the regions $\mathbf{M}_{e}$ of interest of the feature extractor with the regions $\mathbf{M}_{a}$ that are not discarded by the feature aggregator.
Specifically, we align $\mathbf{M}_{a}$ and $\mathbf{M}_{e}$  by Kullback-Leibler divergence~\cite{van2014renyi}:
\begin{equation}
\mathcal{L}_{SA} = \sum^{h\times w}_{i}\mathbf{M}_{a}^{i}log\frac{\mathbf{M}_{e}^{i}}{\mathbf{M}_{a}^{i}}+\sum^{h\times w}_{i}\mathbf{M}_{e}^{i}log\frac{\mathbf{M}_{a}^{i}}{\mathbf{M}_{e}^{i}},
\end{equation}
where $i \in \mathbb{R}^{h\times w}$. In practice, we found that direct alignment leads to network learning difficulties due to the significant difference in distribution between $\mathbf{M}_{a}$ and $\mathbf{M}_{e}$. 
Therefore, we smoothed $\mathbf{M}_{a}$ and $\mathbf{M}_{e}$. Taking $\mathbf{M}_{a}$ as an example, it can be defined as:
\begin{equation}
\mathbf{M}_{a}^{(i)} = 
\begin{cases} 
\mathbf{M}_{a}^{(i)} & \text{if } \mathbf{M}_{a}^{(i)} < \text{perc}_{10}(\mathbf{M}_{a}^{(i)}) \\
\text{perc}_{10}(\mathbf{M}_{a}^{(i)}) & \text{otherwise},
\end{cases}
\end{equation}
where \(\text{perc}_{10}(\mathbf{M}_{a}^{(i)})\) denotes the 10-th percentile value of the \(i\)-th row in matrix \(\mathbf{M}_{a}\). This formula is used to truncate the top 10\% of values in each row to the 10-th percentile value, thereby smoothing the distribution of $\mathbf{M}_{a}$ and $\mathbf{M}_{e}$.

\noindent\textbf{Foreground-Background Contrast Enhancement Loss.}
We further introduce a Foreground-Background Contrast Enhancement Loss (CEL) to encourage the network to generate regions that are both reliable (\textit{i.e.}, produce consistent attention across images) and discriminative (ensure that they are well distinguished from the background).
We first reshape $\mathbf{F}$ containing the patch tokens into $\mathcal{F} \in \mathbb{R}^{h\times w\times d}$ that is consistent with the shape of $M$.
From this, we can obtain the foreground comparison prototype $f_{fg} $:
\begin{equation}
f_{fg} = \frac{\sum^{h\times w}_{i}\mathbf{M}(i)\cdot \mathcal{F}(i)}{||\sum^{H\times W}_{i}\mathbf{M}(i)\cdot \mathcal{F}(i)||},
\label{eq:2}
\end{equation}
where $||\cdot||$ means L2 norm. Similarly, we can get the background prototype $f_{bg}$:
\begin{equation}
f_{bg} = \frac{\sum^{h\times w}_{i}(1-\mathbf{M}(i))\cdot \mathcal{F}(i)}{||\sum^{H\times W}_{i}(1-\mathbf{M}(i))\cdot \mathcal{F}(i)||}.
\label{eq:3}
\end{equation}
Next, Based on triplet loss~\cite{triplet} $\mathcal{L}_{CE}$ can be defined as:
\begin{equation}
 \mathcal{L}_{CE} = \max(0,1-(f_{fg}\cdot f^{p}_{fg}-f_{fg}\cdot f_{bg})),
\end{equation}
where $f^{p}_{fg}$ is the foreground prototype of the positive sample (different images of the same scene), as shown in Figure~\ref{fig:method}. 
Notice that in order to force the network to optimize only $\mathbf{M}$ without affecting feature extraction, we truncate the gradient of $\mathcal{F}$ when computing the $ \mathcal{L}_{CE}$ loss.
Fueled by $ \mathcal{L}_{CE}$, $\mathbf{M}$ tends to produce consistent and reliable foreground discriminative region activation, and foreground and background regions are encouraged to be clearly distinguished.
\begin{algorithm}[t]
\caption{Pseudo-correspondence Ground Truth Construction}
\begin{algorithmic}[1]
\STATE \textbf{Input:} Discriminative Region Map $M$, Assignment Score Matrix $S$
\STATE \textbf{Parameters:} $thr1 = 0.8$, $thr2 = 0.5$, $N=8$
\STATE We sort the patches using the activation values in $M$. Then, we select the patch feature $f_{p}$ with the highest activation value and exclude it from the sequence.
\STATE The corresponding cluster order number is found through the index assignment matrix $S$. Then all the patches in the positive sample image that are assigned to that cluster are found based on the cluster order number as candidate features.
\STATE If candidate features \{{$f_{p^{'}_{1th}}$, $f_{p^{'}_{2th}}$}\} corresponding to $f_{p}$ satisfy
$sim(f_{p},f_{p^{'}_{1th}})>thr1$ and $sim(f_{p},f_{p{'}_{2th}})/sim(f_{p},f_{p^{'}_{1th}})<thr2$ with the highest and second-highest similarity to the corresponding feature $f_{p}$, perform step 6 else perform step 3.
\STATE $p$ and $p^{'}_{1th}$ as pseudo labels $(p,p^{'})$ and then perform step 3 until $N$ labels are obtained or there are no candidates.
\STATE \textbf{Return:} $N$ pseudo-correspondence ground truth.
\end{algorithmic}
\label{alg:algorithm}
\end{algorithm}
With these two losses proposed above, FoL can model reliable discriminative regions and benefit both image retrieval and re-ranking stages.
On the one hand, the network can be induced to pay more attention to critical regions when extracting features and aggregating global features by constraining the attention of the transformer and the score matrix of the aggregation phase. On the other hand, discriminative regions can be used as a priori information in the re-ranking stage to accelerate the efficiency and accuracy of local matching.

\subsection{Weakly Supervised Local Feature Learning}
\label{sec:wsl}
Due to the lack of pixel-level correspondence labels in the VPR dataset, existing two-stage methods can usually only supervise the final matching results, such as R2Former~\cite{R2FormerCVPR2023}'s classification loss and CricaVPR~\cite{crica}'s mutual nearest neighbor loss.
In FoL, we explore the use of the clustering results of local features in the global feature aggregation process to construct pseudo-correspondence truths thereby enabling pixel-level supervision of local feature matching.

\noindent\textbf{Pseudo-correspondence Ground Truth Construction.}
As shown in Figure~\ref{fig:method}, in the global feature aggregation phase, each local patch is assigned to a specific cluster, which in fact indirectly includes the correspondence between patches, \textit{i.e.}, the corresponding pair of patches will be assigned to the same cluster.
We develop the pseudo correspondence truth construction algorithm based on the above observation as shown in Algorithm~\ref{alg:algorithm}.

\noindent\textbf{Weakly Supervised Pseudo correspondence Loss.}
We obtain $(D_{l}^{p}, D_{l}^{p^{'}})$ by sampling the network output dense local features $D_{l}$ using the pseudo correspondence $(p, p^{'})$. From this, $\mathcal{L}_{PC}$ can be defined as:
\begin{equation}
\mathcal{L}_{PC}=\frac{\sum exp(sim(f_{p_{i}},f_{p^{'}_{i}}))\cdot (1-sim(D_{l}^{p_{i}},D_{l}^{p^{'}_{i}})))}{\sum exp(sim(f_{p_{i}},f_{p^{'}_{i}}))}.
\end{equation}
Since not true ground truth is used, we use similarity as a confidence for evaluating the quality of labels to mitigate the negative impact of inaccurate labels on model training, hence $\mathcal{L}_{PC}$ is a weakly supervised loss.

\subsection{Efficient Re-ranking with Discriminative Region Guidance}
\label{sec:erd}
We propose to accelerate and enhance the re-ranking process using the local regions mask $\mathbf{M}$ modeled in previous section as a priori information for local matching.

First, we convert $\mathbf{M}$ into a binary matrix, where the top \(k\) positions in $\mathbf{M}$ are set to 1, and others to 0. This can be expressed as:

\begin{equation}
\mathbf{M}_{\text{bin}} = 
\begin{cases} 
1 & \text{if index} \in \text{top } k \text{ of } \mathbf{M} \\
0 & \text{otherwise},
\end{cases}
\end{equation}
where \(k\) is set to the top $40\%$.

Then, we interpolate \(\mathbf{M}_{\text{bin}}\) and select the parts of the dense local features $D_{l}$ where the mask is 1:

\begin{equation}
D_{l}^{\mathbf{M}} = D_{l}[\mathbf{M}_{\text{bin}} = 1]
\end{equation}

Finally, we only perform nearest-neighbor local feature matching on $D_{l}^{\mathbf{M}}$, which reduces the probability of wrong matching and improves the efficiency of matching due to the narrowed matching range.

\subsection{Total Loss Function}  

Following previous works~\cite{salad}, we also use the multi-similarity loss~\cite{wang2019multi} $\mathcal{L}_{MS}$ with an online hard mining strategy to optimize the global features. 
To optimize the local features matching, we also adopt the mutual nearest neighbor local feature loss \(\mathcal{L}_{MNN}\) proposed in SelaVPR\cite{lu2024towards}. 
Combined with the techniques presented in the previous sections, the total loss can be defined as:
\begin{equation}
\mathcal{L} = \mathcal{L}_{MS}+ \mathcal{L}_{MNN}+ \mathcal{L}_{CE}+ \alpha \mathcal{L}_{SA}+ \beta \mathcal{L}_{PC},
\end{equation}

where $\{\alpha, \beta\}$ are hyperparameters to adjust the magnitude of the different losses.

\section{Experiments}

\subsection{Benchmarks and Performance Evaluation}

Our experiments utilize several VPR benchmark datasets to assess the performance of our models, focusing primarily on Tokyo24/7, Pitts250k, and MSLS, with additional evaluations on Nordland, AmsterTime, SVOX, SPED, and SF-XL. \textbf{Tokyo24/7}~\cite{densevlad} includes approximately 76,000 database images and 315 query images from urban environments, each image involving three different viewpoints and three different times of the day, showcasing significant changes in lighting conditions. \textbf{Pitts250k}~\cite{torii2013visual} consists of images from Google Street View panoramas, exhibiting significant viewpoint changes, moderate condition variations, and a small number of dynamic objects. \textbf{MSLS} (Mapillary Street-Level Sequences)~\cite{warburg2020mapillary} is a large-scale dataset with over 1.6 million images collected from urban, suburban, and natural scenes, testing models under varying conditions such as illumination, weather, and dynamic objects. \textbf{Nordland}~\cite{sunderhauf2013we} captures images from a fixed viewpoint in the front of a train on the same route over four seasons, providing images from suburban and natural environments with significant seasonal and lighting changes. \textbf{AmsterTime}~\cite{yildiz2022amstertime} presents historical grayscale queries and contemporary RGB references from Amsterdam, challenging models with temporal and modality variations. \textbf{SVOX}~\cite{Berton_2021_svox} evaluates performance under different weather and lighting conditions, with queries extracted from the Oxford RobotCar dataset. The \textbf{SPED}~\cite{zaffar2021vpr} and \textbf{SF-XL}~\cite{cosplace} datasets add further diversity with varying degrees of viewpoint and condition changes. We follow the common evaluation metrics used in previous works~\cite{ali2024boq}. We assess performance using Recall@N, with a 25-meter threshold for Tokyo24/7, Pitts30k, and MSLS, and $\pm 10$ frames for Nordland, effectively measuring retrieval accuracy under various conditions.

\begin{table*}[t]
\setlength{\extrarowheight}{0pt}
\setlength{\aboverulesep}{0pt}
\setlength{\belowrulesep}{0pt}
\small
\centering
\setlength{\tabcolsep}{1.4pt} 
\renewcommand{\arraystretch}{1} 
{
\begin{tabular}{@{}l||ccc||ccc||ccc||ccc}
\toprule[1.5px]
\multirow{2}{*}{Method} & \multicolumn{3}{c||}{Pitts250k-test} & \multicolumn{3}{c||}{MSLS-val} & \multicolumn{3}{c||}{MSLS-challenge} & \multicolumn{3}{c}{Tokyo24/7} \\
\cline{2-13}
& R@1 & R@5 & R@10 & R@1 & R@5 & R@10 & R@1 & R@5 & R@10 & R@1 & R@5 & R@10 \\
\hline
NetVLAD~$_{\textcolor{blue}{\text{CVPR' 2016}}}$ \cite{netvlad} & 90.5 & 96.2 & 97.4 & 53.1 & 66.5 & 71.1 & 35.1 & 47.4 & 51.7 & 60.6 & 68.9 & 74.6 \\
SFRS~$_{\textcolor{blue}{\text{ECCV' 2020}}}$ \cite{sfrs} & 90.7 & 96.4 & 97.6 & 69.2 & 80.3 & 83.1 & 41.6 & 52.0 & 56.3 & 81.0 & 88.3 & 92.4 \\
Patch-NetVLAD~$_{\textcolor{blue}{\text{CVPR' 2021}}}$ \cite{patchvlad} $^\dag$ & - & - & - & 79.5 & 86.2 & 87.7 & 48.1 & 57.6 & 60.5 & 86.0 & 88.6 & 90.5 \\
CosPlace~$_{\textcolor{blue}{\text{CVPR' 2022}}}$ \cite{cosplace} & 92.4 & 97.2 & 98.1 & 82.8 & 89.7 & 92.0 & 61.4 & 72.0 & 76.6 & 81.9 & 90.2 & 92.7 \\
MixVPR~$_{\textcolor{blue}{\text{WACV' 2023}}}$ \cite{mixvpr} & 94.6 & 98.3 & 99.0 & 88.2 & 93.1 & 94.3 & 64.0 & 75.9 & 80.6 & 86.7 & 92.1 & 94.0 \\
R2Former~$_{\textcolor{blue}{\text{CVPR' 2023}}}$ \cite{R2FormerCVPR2023} $^\dag$ & 93.2 & 97.5 & 98.3 & 89.7 & 95.0 & 96.2 & 73.0 & 85.9 & 88.8 & 88.6 & 91.4 & 91.7 \\
EigenPlaces~$_{\textcolor{blue}{\text{ICCV' 2023}}}$ \cite{eigenplaces} & 94.1 & 98.0 & 98.7 & 89.1 & 93.8 & 95.0 & 67.4 & 77.1 & 81.7 & 93.0 & 96.2 & 97.5 \\
SelaVPR~$_{\textcolor{blue}{\text{ICLR' 2024}}}$ \cite{sela} $^\dag$ & 95.7 & 98.8 & \underline{99.2} & 90.8 & \underline{96.4} & {97.2} & 73.5 & 87.5 & 90.6 & 94.0 & 96.8 & 97.5 \\
CricaVPR~$_{\textcolor{blue}{\text{CVPR' 2024}}}$ \cite{crica} & 95.6 & 98.9 & \textbf{99.5} & 90.0 & 95.4 & 96.4 & 69.0 & 82.1 & 85.7 & 93.0 & 97.5 & 98.1 \\
SALAD~$_{\textcolor{blue}{\text{CVPR' 2024}}}$ \cite{salad} & 95.1 & 98.5 & 99.1 & {92.2} & 96.2 & {97.0} & 75.0 & 88.8 & \underline{91.3} & 94.6 & 97.5 & 97.8 \\
BoQ~$_{\textcolor{blue}{\text{CVPR' 2024}}}$ \cite{ali2024boq} & 95.0 & 98.4 & 99.1 & 91.4 & 94.5 & 96.1 & - & - & - & - & - & - \\
\rowcolor{lightblue}
FoL-global & {\underline{96.5}} & {\underline{99.1}} & {\textbf{99.5}} & {\underline{93.1}} & {\textbf{96.9}} & {\underline{97.4}} & \underline{78.7} & {\underline{90.8}} & {\textbf{93.0}} & \underline{96.2} & {\underline{98.7}} & {\underline{98.7}} \\
\rowcolor{lightblue}
FoL-re-ranking$^\dag$ & {\textbf{97.0}} & {\textbf{99.2}} & {\textbf{99.5}} & {\textbf{93.5}} & {\textbf{96.9}} & {\textbf{97.6}} & {\textbf{80.0}} & {\textbf{90.9}} & {\textbf{93.0}} & {\textbf{98.4}} & {\textbf{99.1}} & {\textbf{99.4}} \\
\bottomrule[1.5px]
\end{tabular}}
\caption{Comparison to state-of-the-art methods on four benchmarks. The best results are highlighted in \textbf{bold} and the second best are \underline{underlined}. Two-stage methods are marked with \dag.}
\label{tab:compare_SOTA}
\end{table*}
\begin{table*}[h]
\setlength{\extrarowheight}{0pt}
\setlength{\aboverulesep}{0pt}
\setlength{\belowrulesep}{0pt}
\small
\centering
\setlength{\tabcolsep}{0.9pt} 
\renewcommand{\arraystretch}{1} 
\begin{tabular}{@{}lccccccccccc}
\toprule[1.5px]
\multicolumn{1}{c}{Method} & \multicolumn{1}{c}{Nordland} & \multicolumn{1}{c}{\begin{tabular}[c]{@{}c@{}}Amster\\ Time\end{tabular}} & \multicolumn{1}{c}{\begin{tabular}[c]{@{}c@{}}SF-XL\\ Occlusion\end{tabular}} & \multicolumn{1}{c}{\begin{tabular}[c]{@{}c@{}}SF-XL\\ Night\end{tabular}} & \multicolumn{1}{c}{\begin{tabular}[c]{@{}c@{}}SVOX\\ Night\end{tabular}} & \multicolumn{1}{c}{\begin{tabular}[c]{@{}c@{}}SVOX\\ Sun\end{tabular}} & \multicolumn{1}{c}{\begin{tabular}[c]{@{}c@{}}SVOX\\ Snow\end{tabular}} & \multicolumn{1}{c}{\begin{tabular}[c]{@{}c@{}}SVOX\\ Overcast\end{tabular}} & \multicolumn{1}{c}{SVOX} & \multicolumn{1}{c}{SPED} \\ \midrule
SFRS~$_{\textcolor{blue}{\text{ECCV' 2020}}}$ \cite{sfrs} & 16.0 & 29.7 & - & - & 28.6 & 54.8 & - & - & - & - \\
CosPlace~$_{\textcolor{blue}{\text{CVPR' 2022}}}$ \cite{cosplace} & 43.8 & 38.7 & - & - & 44.8 & 67.3 & - & - & - & - \\
MixVPR~$_{\textcolor{blue}{\text{WACV' 2023}}}$ \cite{mixvpr} & 58.4 & 40.2 & - & - & 64.4 & 84.8 & - & - & - & - \\
EigenPlaces~$_{\textcolor{blue}{\text{ICCV' 2023}}}$ \cite{eigenplaces} & 54.4 & 48.9 & 32.9 & 23.6 & 58.9 & 86.4 & 93.1 & 93.1 & 98.0 & 70.2 \\
SelaVPR~$_{\textcolor{blue}{\text{ICLR' 2024}}}$ \cite{sela} & 85.2 & 55.2 & 35.5 & 38.4 & 89.4 & 90.2 & 97.0 & 97.0 & 97.2 & 88.6 \\
CricaVPR~$_{\textcolor{blue}{\text{CVPR' 2024}}}$ \cite{crica} & \underline{90.7} & \underline{64.7} & 42.1 & 35.4 & 85.1 & 93.8 & 96.0 & 96.7 & 97.8 & 91.3 \\
SALAD~$_{\textcolor{blue}{\text{CVPR' 2024}}}$ \cite{salad} & 76.0 & 58.8 & \underline{51.3} & 46.6 & 95.4 & 97.2 & {98.9} & \textbf{98.3} & 98.2 & \textbf{92.1} \\
BoQ~$_{\textcolor{blue}{\text{CVPR' 2024}}}$ \cite{ali2024boq} & 74.4 & 53.0 & - & - & 85.2 & 96.5 & 98.4 & \textbf{98.3} & - & 86.2 \\

\rowcolor{lightblue}
FoL-global & 87.8 & 64.6 & \underline{51.3} & \underline{53.4} & \underline{98.3} & \underline{98.1} & \underline{99.1} & \underline{97.9} & \underline{98.4} & {\textbf{92.1}} \\ 
\rowcolor{lightblue}
FoL-re-ranking & \textbf{92.6} & \textbf{70.1} & \textbf{61.8} & {\textbf{60.5}} & \textbf{98.8} & \textbf{98.8} & \textbf{99.3} & {\textbf{98.3}} & \textbf{98.9} & \underline{91.8} \\ 

\bottomrule[1.5px]
\end{tabular}
\caption{Comparison (R@1) to state-of-the-art methods on more challenging datasets.
}  
\label{tab:special_cases_results}
\end{table*}

\subsection{Implementation Details}

We initialize the ViT-L backbone with pre-trained DINOv2 weights and fine-tune only the last four layers of the backbone. The remaining modules are set to learnable. In the feature extraction stage, the number of clusters $M$ is set to 64. In the re-ranking stage, the number of channels up-conv are $256$ and $128$, and the convolution kernel size was 3$\times$3, stride=2, padding=1. 
The ViT architecture supports variable input sizes, provided images can be partitioned into 14 × 14 patches. To speed up training, we used 322 × 322 images but evaluated on 504 × 504 resolution.
We trained with one A100 GPU on GSV-Cities~\cite{gsv}, a large city location dataset collected by Google Street View. Batch size is $60$ and each batch is described by $4$ images. The AdamW optimizer with a linear learning rate schedule was used, with a learning rate of $6e-5$ and a weight decay of $9.5e-9$. The training converged after $5$ epochs.
To ensure the validity of our experiments and optimize hyperparameter selection, we continuously monitored recall performance on the MSLS validation set. We follow mainstream works to use 25 meters as the threshold for correct scene and report recall@k (k=1,5,10) as evaluation metrics.

\subsection{Comparisons with state-of-the-art Methods}

In this section, we compare our one-stage (FoL-global) and two-stage (FoL-re-ranking) results with previous state-of-the-art VPR methods, including eight one-stage methods that utilize global feature retrieval: NetVLAD \cite{netvlad}, SFRS \cite{sfrs}, CosPlace \cite{cosplace}, MixVPR \cite{mixvpr}, EigenPlaces \cite{eigenplaces}, CricaVPR \cite{crica}, SALAD \cite{salad}, and BoQ \cite{ali2024boq}. Additionally, we compare our method with three two-stage methods that incorporate re-ranking: Patch-NetVLAD \cite{patchvlad}, R2Former \cite{R2FormerCVPR2023}, and SelaVPR \cite{sela}. Our method, similar to CricaVPR and SALAD, is trained on the GSV-Cities~\cite{gsv}. Quantitative results are presented in Table \ref{tab:compare_SOTA}.
Our FoL-global method achieves the best R@1/R@5/R@10 across all datasets using only global features for direct retrieval, significantly outperforming other one-stage and two-stage methods. Specifically, FoL-global improves the absolute R@1 by 0.8\%, 0.9\%, 3.7\%, and 1.6\% on Pitts250k-test, MSLS-val, MSLS-challenge, and Tokyo24/7, respectively. This highlights the effectiveness of the single global representation learned through our approach.
After re-ranking with local features, the complete FoL method outperforms all other methods by a substantial margin. Specifically, FoL-re-ranking improves the absolute R@1 by 1.3\%, 1.3\%, 5.0\%, and 3.8\% on Pitts250k-test, MSLS-val, MSLS-challenge, and Tokyo24/7, respectively. These results demonstrate that our model can overcome various visual changes induced by illumination (Tokyo24/7), viewpoint (Pitts250k), weather, season, and dynamic objects (MSLS).

Subsequently, we demonstrate the effectiveness of our method in the most challenging VPR benchmark scenarios through experiments on ten demanding datasets. The detailed results in Table \ref{tab:special_cases_results} emphasize the significant advantages of our method over previous methods on these datasets. FoL-re-ranking shows improvements of +1.9\%, +5.4\%, +10.5\%, +13.9\%, +3.4\%, and +1.6\% on Nordland, Amster Time, SF-XL Occlusion, SF-XL Night, SVOX Night, and SVOX Sun, respectively. Notably, we observe an improvement of over 10\% on the large-scale SF-XL dataset, demonstrating that our model can generalize to highly challenging scenarios involving viewpoint changes and occlusions.

\begin{table}[H]
    \centering
    \setlength{\tabcolsep}{2.5mm}{
    \renewcommand\arraystretch{0.8} {
    \begin{tabular}{lccc}
    \toprule[1.5px]
    Method & \begin{tabular}[c]{@{}c@{}}Extraction\\ Time (s)\end{tabular} &         \begin{tabular}[c]{@{}c@{}}Matching\\ Time (s)\end{tabular} & \begin{tabular}[c]{@{}c@{}}Total \\ Time (s)\end{tabular} \\
    
    \midrule
   
    TransVPR & 0.007 & 2.632 & 2.639 \\
    R2Former & 0.009 & 0.347 & 0.356 \\
    SelaVPR & \textbf{0.023} & 0.071 & 0.094 \\ \rowcolor{lightblue}
    \midrule
    FoL-re-ranking  & 0.025 & \textbf{0.032} & \textbf{0.057} \\
    \bottomrule[1.5px]
    \end{tabular}}
    }
     \caption{The single query runtime comparison of two-stage methods on Pitts250k-test.}
    \label{tab:time}
\end{table}
\begin{figure}[t]
    \centering
    \includegraphics[width=1\linewidth]{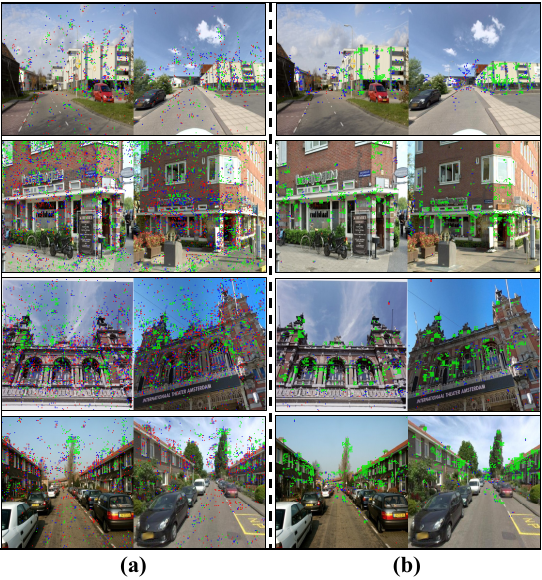}
    \caption{Visualization of local feature matching in the re-ranking stage. (a) shows the \textit{w/o} Discriminative Region Guidance, while (b) displays \textit{w/} Discriminative Region Guidance. Red $\color{red}\bullet$, Blue $\color{blue}\bullet$, and Green $\color{green}\bullet$ represent low, medium, and high similarity matching points, respectively (${ \color{red}\bullet} \rightarrow { \color{blue}\bullet} \rightarrow { \color{green}\bullet}$ indicates increasing similarity).
}
    \label{fig:Matching_demo}
\end{figure}

\begin{figure}[t]
    \centering
    \includegraphics[width=1\linewidth]{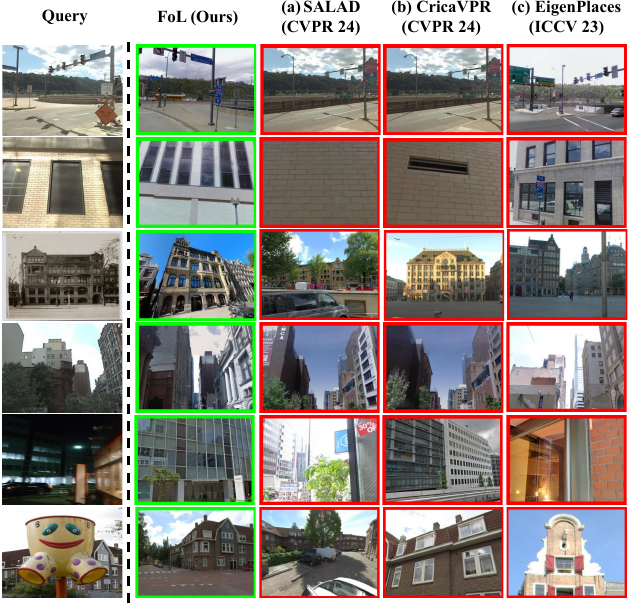}
    \caption{Qualitative VPR comparison results. 
    Our FoL accurately matching the query images, while other methods such as SALAD, CricaVPR, and EigenPlaces cause erroneous matches under complex lighting and viewpoint variations.
}
\label{fig:Matching_demo2}
\vspace{-0.3cm}
\end{figure}

Efficiency is another critical metric for evaluating VPR methods. In Table \ref{tab:time}, we compare the runtime of our method with other two-stage methods on Pitts250k-test, including both feature extraction time and matching/retrieval time. Due to the absence of geometric verification in our re-ranking stage and the involvement of only discriminative regions in matching, our method demonstrates a significant speed advantage.
Figure~\ref{fig:Matching_demo} and Figure~\ref{fig:Matching_demo2} show the visualization results of re-ranking local matching and  VPR retrieval.

\subsection{Ablation Study}

Table~\ref{table:ab} reports the results of our proposed FoL ablation experiments on the MSLS Challenge benchmark.
Here, the Baseline denotes using only the SALAD model trained with $\mathcal{L}_{MS}$ and $\mathcal{L}_{MNN}$ losses, without any of the proposed techniques.
We first apply a simple re-ranking strategy to the Baseline, which aims to refine the initial retrieval results by re-examining candidate lists. This step slightly improves performance from (74.8, 88.4, 91.1) to (75.7, 89.1, 91.8).
Next, we introduce $\mathcal{L}_{SA}$ to explicitly model discriminative regions for the VPR task. After applying $\mathcal{L}_{SA}$, we observe increments of ($+1.5\%$, $+0.3\%$, $+0.9\%$) on R@1, R@5, and R@10, respectively. Subsequently, adding $\mathcal{L}_{CE}$ further leverages spatial information, yielding increments of ($+0.9\%$, $+1.2\%$, $+0.1\%$).
Then, we impose $\mathcal{L}_{PC}$ loss to enhance local feature representation in the re-ranking phase. As illustrated in Table~\ref{table:ab}, this leads to another improvement of ($+0.6\%$, $+0.2\%$, $+0.2\%$), highlighting the crucial role of improving local feature quality for re-ranking.
Finally, we adopt our proposed Efficient Re-ranking with Discriminative Region Guidance \textit{i.e.}, DRG. The performance reaches (80.0, 90.9, 93.0), and notably, the proposed DRG significantly reduces the matching time compared to dense local feature matching, resulting in a substantial inference speedup.

\begin{table}[h]
\setlength{\extrarowheight}{0pt}
\setlength{\aboverulesep}{0pt}
\setlength{\belowrulesep}{0pt}
\centering
\setlength{\tabcolsep}{3mm}{
\renewcommand\arraystretch{1.2} {
\begin{tabular}{l|ccc} 
\toprule[1.5px]
\multirow{2}{*}{Configurations} & \multicolumn{3}{c}{MSLS Challenge}  \\
                         & R@1 & R@5 & R@10                    \\ 
\hline
Baseline                 & 74.8    & 88.4    & 91.1                                              \\ 
\hdashline
$+ \text{re-ranking}$                     & 75.7    & 89.1    & 91.8                                           \\
$+ \mathcal{L}_{SA}$                     & 77.2   & 89.4   & 92.7                                           \\
$+\mathcal{L}_{CE}$                      & 78.1   & {90.6}   & \underline{92.8}                                          \\ 
$+\mathcal{L}_{PC}$        & \underline{78.7} & {\underline{90.8}} & {\textbf{93.0}}                                           \\ 
\hdashline \rowcolor{lightblue}
$+$ DRG                     & {\textbf{80.0}} & {\textbf{90.9}} & {\textbf{93.0}}                                   \\
\bottomrule[1.5px]
\end{tabular}
}}
\caption{Ablation Study on MSLS Challenge benchmark. Here DRG means Efficient Re-ranking with Discriminative Region Guidance.
}  
\label{table:ab}
\vspace{-0.2cm}
\end{table}

\section{Conclusions}
In this paper, we propose a two-stage VPR method called FoL and focus our research vision on the role of mining spatial local information for VPR.
We explicitly model reliable and discriminative local regions with two well-designed losses and demonstrate the importance of discriminative region modeling for image retrieval and local matching in VPR tasks.
We also introduce a weakly supervised local feature training strategy based on pseudo-correspondence to improve re-ranking performance.
In addition, we propose discriminative region-guided efficient re-ranking strategy to further improve the accuracy and efficiency.
Experimental results show that our FoL achieves the latest state-of-the-art performance on mainstream benchmarks for VPR tasks with high efficiency.

\section{Acknowledgements}
This work was supported by National Science and Technology Major Project (No.2022ZD0116800), and in part by the Taishan Scholars Program (Nos. TSQN202211214 and TSQN202408245), Shandong Excellent Young Scientists Fund Program (Overseas) No.2023HWYQ-113, Beijing Natural Science Foundation (No.JQ23014), National Natural Science Foundation of China (Nos.62271074, 32271983, 62171321, 62162044), Shandong Provincial Natural Science Foundation for Young Scholars (Nos.ZR2024QF110 and ZR2024QF052) and the Open Project Program of State Key Laboratory of Virtual Reality Technology and Systems, Beihang University (No.VRLAB2023B01). 

\bibliography{aaai25}

\newpage

\appendix

\twocolumn[{%
 \centering
 \LARGE [Supplementary Material] \\ Focus on Local: Finding Reliable Discriminative Regions for Visual Place Recognition\\[1.5em]
}]

In this supplementary material, we present a concise overview of the architecture and training strategies behind our proposed FoL method. FoL is built upon a DINOv2 pre-trained ViT-L backbone, which encodes image patches into 1024-dimensional features, with a learnable [cls] token aggregating global context. A NetVLAD-style learnable aggregator, enhanced by a Sinkhorn-based assignment, is used to cluster features and construct global descriptors. We also describe the generation of discriminative region masks by combining attention-based and cluster-based cues, enabling robust retrieval and re-ranking. Additional implementation details, such as preprocessing steps and training on GSV-Cities, are provided. Finally, we include qualitative results, ablation studies, and extensive evaluations across diverse benchmarks to demonstrate the effectiveness and generalizability of FoL. 

\textbf{Code and pretrained models} are available at \url{https://github.com/chenshunpeng/FoL}.


\subsection{More Implementation Details}

\noindent\textbf{Feature Extractor:}

Following prior works~\cite{keetha2023anyloc,sela,salad,crica}, we adopt the pre-trained DINOv2 model~\cite{oquab2023dinov2} as the feature extractor for our visual place recognition (VPR) pipeline. Specifically, we utilize the ViT-L configuration, which incorporates a greater number of transformer layers and a higher embedding dimensionality compared to smaller ViT variants. In ViT-L, each image is divided into non-overlapping patches that are projected into a 1024-dimensional embedding space. A special [CLS] token is prepended to these patch embeddings, forming a sequence that includes a global representation of the image.
The resulting sequence is processed by a stack of transformer blocks, producing a high-dimensional feature representation. This feature is particularly well-suited for tasks that require fine-grained semantic understanding, such as VPR, due to its ability to encode rich and detailed image information. Below, we briefly describe the ViT processing pipeline.

An input image is initially partitioned into \( N \) non-overlapping patches, each of which is mapped to a \( D \)-dimensional embedding, yielding \( \mathbf{x}_p \in \mathbb{R}^{N \times D} \). A learnable class token \( \mathbf{x}_{\text{cls}} \) is prepended to form the sequence \( \mathbf{x}_0 = [\mathbf{x}_{\text{cls}}; \mathbf{x}_p] \in \mathbb{R}^{(N+1) \times D} \). Positional embeddings are added to this sequence to preserve spatial structure, and the resulting sequence \( \mathbf{z}_0 \) is passed through a series of transformer layers.
Each transformer block consists of a multi-head attention (MHA) mechanism, a multi-layer perceptron (MLP), and layer normalization (LN). The forward computation at the \( l \)-th block is defined as:

\[
\mathbf{z}_l' = \text{MHA}(\text{LN}(\mathbf{z}_{l-1})) + \mathbf{z}_{l-1},
\]
\[
\mathbf{z}_l = \text{MLP}(\text{LN}(\mathbf{z}_l')) + \mathbf{z}_l',
\]
where \( \mathbf{z}_{l-1} \) and \( \mathbf{z}_l \) denote the input and output of the \( l \)-th transformer layer, respectively.
The MLP module typically comprises two fully connected layers and non-linear activation functions, facilitating feature transformation. The MHA module captures dependencies across different spatial locations by computing the scaled dot-product attention:

\[
\text{Attn}(\mathbf{Q}, \mathbf{K}, \mathbf{V}) = \text{Softmax}\left(\frac{\mathbf{QK}^T}{\sqrt{d}}\right) \mathbf{V},
\]
where \( \mathbf{Q} \), \( \mathbf{K} \), and \( \mathbf{V} \) are the query, key, and value matrices, respectively, and \( d \) is the dimension of the key vectors.

\noindent\textbf{Learnable Aggregation:}

To generate global image descriptors, we employ a NetVLAD-style learnable aggregator proposed in SALAD~\cite{salad}, which builds upon the classic NetVLAD~\cite{netvlad} framework. In VPR tasks, the [CLS] token is commonly used as the global representation, while the remaining tokens encode local features. SALAD extends this approach by performing a differentiable clustering of local features using an optimal transport mechanism based on the Sinkhorn algorithm, and subsequently concatenates the aggregated local features with the global [CLS] token. Below, we describe the key components of the SALAD aggregator.

\textbf{Feature Assignment.}
Let $\mathbf{f}_i$ denote the local feature derived from token $\mathbf{t}_i$. A score matrix $\mathbf{S} \in \mathbb{R}_{>0}^{n \times m}$ is computed to evaluate the similarity between features and learnable cluster centers. Specifically, the unnormalized affinity between feature $\mathbf{f}_i$ and the $j$-th cluster is defined as:

\begin{equation}
\label{eq:Sinkhorn}
P_{i,j} = \text{Sinkhorn}\left(\exp(\mathbf{w}_j^\top \mathbf{f}_i + b_j)\right),
\end{equation}
where $\mathbf{w}_j$ and $b_j$ are the learnable weights and biases for cluster $j$. The Sinkhorn algorithm is applied to normalize the score matrix $\mathbf{S}$ into an assignment matrix $\mathbf{P}$ that satisfies the marginal constraints: $\mathbf{P} \mathbf{1}_{m+1} = \boldsymbol{\mu}$ and $\mathbf{P}^\top \mathbf{1}_n = \boldsymbol{\kappa}$, ensuring a balanced feature-to-cluster distribution. After normalization, the dustbin column (representing unassigned features) is discarded:

\begin{equation}
\label{eq:dustbin}
\mathbf{P} = \left[ \mathbf{p}_{\ast,1}, \dots, \mathbf{p}_{\ast,m} \right],
\end{equation}
where each column $\mathbf{p}_{\ast,j}$ represents the soft assignment of all features to the $j$-th cluster.

\textbf{Feature Aggregation.}
Using the assignment matrix $\mathbf{P}$, features are aggregated to form a cluster-wise descriptor matrix $\mathbf{V}$. Each element $V_{j,k}$ of the matrix is computed by weighted summation over the corresponding feature dimension:

\begin{equation}
V_{j,k} = \sum_{i=1}^{n} P_{i, j} \cdot f_{i,k},
\end{equation}
where $f_{i,k}$ denotes the $k$-th dimension of the dimensionally-reduced feature vector $\mathbf{f}_i$. These local features are typically projected into a lower-dimensional space prior to assignment.
A global scene representation $\mathbf{g}$ is extracted from the [CLS] token through a learnable projection. The final global descriptor $\mathbf{D}_{\text{g}}$ is then obtained by concatenating the global vector $\mathbf{g}$ with the flattened and intra-normalized cluster descriptor $\mathbf{V}$, followed by L2 normalization:

\begin{equation}
\mathbf{D}_{\text{g}} = \text{L2Norm}([\mathbf{g}; \text{L2Norm}(\mathbf{V})]),
\end{equation}
where $\mathbf{D}_{\text{g}}$ serves as the final representation of the input scene, encoding both global semantics and spatially-discriminative local features.

\noindent\textbf{The Generation Process of Discriminative Region:}

\textbf{For attention mask $\mathbf{M}_{e}$}, the self-attention score computation of the transformer architecture incorporates the level  of attention paid to each token representing an image  patch, which can be viewed as modeling spatial information.  Specifically, $\mathbf{M}_{e}$ is obtained by averaging the attention  scores of the last multi-head attention layer [cls] token over patch tokens.  
For $\mathbf{M}_{e}$, image patches are processed through transformer blocks using self-attention to capture patterns and contextual information. The self-attention mechanism derives the query, key, and value matrices from the normalized input. After calculating the attention matrix, we select the attention scores from the first  [cls] token to all patch tokens and average over all attention heads to construct $\mathbf{M}_{e}$:

\begin{equation}
\mathbf{M}_{e} = \frac{1}{h} \sum_{i=1}^{h} \text{Attn}(\mathbf{Q}_{\text{cls}}, \mathbf{K}, \mathbf{V})
\end{equation}
\( h \) denote the number of attention heads, $\mathbf{M}_{e}$ represent the average attention values from the class token to each patch.

\textbf{For discriminative foreground region mask $\mathbf{M}_{a}$}, the dustbin cluster  is dedicated to hold background features, while other foreground features can be obtained by assigning to the set of $M$ clusters, thus we can obtain discriminative foreground region mask $\mathbf{M}_{a}$.
For $\mathbf{M}_{a}$, after obtaining \( P_{i,j} \) through Eq. (\ref{eq:Sinkhorn}) during the feature extraction stage, we construct $\mathbf{M}_{a}$ using cluster-based selection. First, the Sinkhorn algorithm is applied to the score matrix to obtain the assignment matrix \(\mathbf{P}\), effectively distributing feature mass across clusters by iteratively normalizing its rows and columns. To refine the assignments and discard irrelevant features, we remove the dustbin column, as described in Eq. (\ref{eq:dustbin}). Then, \(\mathbf{M}_{a}\) is obtained by first averaging the assignment matrix \(\mathbf{P}\) across all features, and then applying a softmax function to normalize the resulting values. This yields a probability distribution that reflects the relative importance of each cluster, which is subsequently used for feature selection and refinement.

Finally, we obtain the discriminative region mask $\mathbf{M}$ by fusing $\mathbf{M}_{e}$ and $\mathbf{M}_{a}$, \textit{i.e.}, $\mathbf{M}=(\mathbf{M}_{e}+\mathbf{M}_{a})/2$.

\noindent\textbf{Training and Preprocessing:}

\textbf{GSV-Cities.} GSV-Cities~\cite{gsv} is a large urban location dataset collected via Google Street View. During training, images are resized to 322x322 pixels using bilinear interpolation, followed by three random augmentations to increase variability. The images are then converted to tensors and normalized according to the dataset's mean and standard deviation.

\noindent\textbf{Evaluation  Dataset:}

\textbf{Pittsburgh}~\cite{torii2013visual}.The Pittsburgh dataset comprises 24 images from different viewpoints at each location, sourced from Google Street View panoramas. It features significant viewpoint changes and moderate condition variations.

\textbf{Mapillary Street-Level Sequences (MSLS)}~\cite{warburg2020mapillary}. MSLS is a large-scale dataset with over 1.6 million images tagged with GPS coordinates and compass angles, captured across 30 cities over seven years. It includes various challenging visual conditions and is split into training, validation (MSLS-val), and test (MSLS-challenge) sets. We use MSLS-val and MSLS-challenge for evaluation.

\textbf{Nordland}~\cite{sunderhauf2013we}. The Nordland dataset contains images from suburban and natural environments, captured from a consistent train viewpoint across four seasons. The dataset focuses on severe condition changes but no viewpoint variations, with frame-level correspondence for ground truth. 

\textbf{Tokyo24/7}~\cite{densevlad}. Tokyo24/7 includes 75,984 database images and 315 query images from urban scenes. Query images are taken from 1,125 images at 125 locations, covering three viewpoints and times of day. This dataset presents challenges like significant viewpoint and day-night variations.

\textbf{St. Lucia}~\cite{glover2010fab}. The St. Lucia dataset consists of ten video sequences from a suburban road in Brisbane. Following the visual geo-localization benchmark, we use the first and last sequences as reference and query data, selecting one image every 5 meters, resulting in 1,549 and 1,464 images, respectively.

\textbf{Eynsham}~\cite{cummins2009highly} is a grayscale image dataset captured by a car-mounted street-view camera along the same Oxford countryside route, recorded twice. It contains 23,935 queries and 23,935 gallery images, with the lack of color adding complexity.

\textbf{SVOX}~\cite{Berton_2021_svox} assesses performance across various weather and lighting conditions, utilizing query sets extracted from the Oxford RobotCar dataset. These queries encompass diverse weather scenarios, including night, overcast, rainy, snowy, and sunny conditions.

\begin{table*}[t]
\centering
\begin{tabular}{@{}lccc|ccc|ccc|ccc@{}}
\toprule[1.5px]
& \multicolumn{3}{c|}{Pitts30k} & \multicolumn{3}{c|}{Eynsham} & \multicolumn{3}{c|}{SVOX Rain} & \multicolumn{3}{c}{St. Lucia} \\
Method & R@1 & R@5 & R@10 & R@1 & R@5 & R@10 & R@1 & R@5 & R@10 & R@1 & R@5 & R@10 \\ \midrule
EigenPlaces & 92.5 & 96.8 & 97.6 & 90.7 & 94.4 & 95.4 & 90.0 & 96.4 & 98.0 & 99.6 & \underline{99.9} & \textbf{100.0} \\
SelaVPR & 92.8 & 96.8 & 97.7 & \multicolumn{1}{c}{90.6} & \underline{95.3} & \underline{96.2} & 94.7 & 98.5 & 99.1 & 99.8 & \textbf{100.0} & \textbf{100.0} \\
CricaVPR & \textbf{94.9} & \underline{97.3} & \textbf{98.2} & 91.6 & 95.0 & 95.8 & 94.8 & 98.5 & 98.7 & \underline{99.9} & \underline{99.9} & \underline{99.9} \\
SALAD & 92.4 & 96.3 & 97.4 & 91.6 & 95.1 & 95.9 & \textbf{98.5} & \underline{99.7} & \textbf{99.9} & \textbf{100.0} & \textbf{100.0} & \textbf{100.0} \\
\midrule
\rowcolor{lightblue}
FoL-global & 93.9 & 97.2 & \underline{98.1} & \underline{91.7} & \underline{95.3} & \underline{96.2} & 96.5 & 99.6 & \underline{99.7} & \underline{99.9} & \textbf{100.0} & \textbf{100.0} \\ 
\rowcolor{lightblue}
FoL-re-ranking & \underline{94.5} & \textbf{97.4} & \textbf{98.2} & {\textbf{92.4}} & \textbf{95.8} & \textbf{96.6} & \underline{98.2} & \textbf{99.9} & \textbf{99.9} & \underline{99.9} & \textbf{100.0} & \textbf{100.0} \\ 
\bottomrule[1.5px]
\end{tabular}
\caption{Comparison to state-of-the-art on More Datasets}
\label{tab:all_results_3}
\vspace{-0.5cm}
\end{table*}

\textbf{AmsterTime}~\cite{yildiz2022amstertime} is a challenging dataset of 1,231 image pairs from Amsterdam, each containing a historical grayscale query and a contemporary RGB reference of the same location. It introduces significant variations in time, modality, viewpoint, and camera characteristics, capturing long-term urban changes.

\textbf{SPED}~\cite{zaffar2021vpr} is a dataset consisting of 607 query images and 607 reference images, all captured by surveillance cameras over time. This dataset introduces significant challenges for VPR due to the substantial variations in viewpoint, as well as seasonal and illumination changes, making it a robust benchmark for evaluating VPR models.

\textbf{SF-XL}~\cite{cosplace} is a large-scale dataset of San Francisco, featuring 360° panoramas, 7,983 queries, and 2.8 million test gallery images. It includes four query sets covering diverse challenges: viewpoint and lighting variations (v1), landmark recognition (v2), night scenes, and heavy occlusion.

\noindent\textbf{Training and Evaluation:}

The ViT-L backbone is initialized with pre-trained DINOv2 weights, with fine-tuning applied only to the last four layers, while the remaining layers remain learnable. During feature extraction, the number of clusters \(M\) is set to 64, and a multi-layer perceptron (MLP) is used to compress the feature dimensions from 1024 to 128. In the re-ranking stage, we use up-convolution layers with 256 and 128 channels, employing \(3 \times 3\) kernels with stride 2 and padding 1. The model is trained on a single A100 GPU at a resolution of \(322 \times 322\) using the GSV-Cities dataset and tested at \(504 \times 504\) resolution. Following common practice, we adopt a 25-meter threshold for determining correct scene retrieval and report recall@k (k=1, 5, 10) as our evaluation metric.

\subsection{More Challenging Datasets}

To ensure comprehensiveness, Table~\ref{tab:all_results_3} below presents the results on datasets not shown in the main paper. Specifically, these datasets include \textbf{Pitts30k}, \textbf{Eynsham}, \textbf{SVOX Rain}, and \textbf{St. Lucia}. Our method, FoL, demonstrates state-of-the-art performance on the majority of these datasets.

\begin{figure}[!h]
    \centering
    \includegraphics[width=0.5\textwidth]{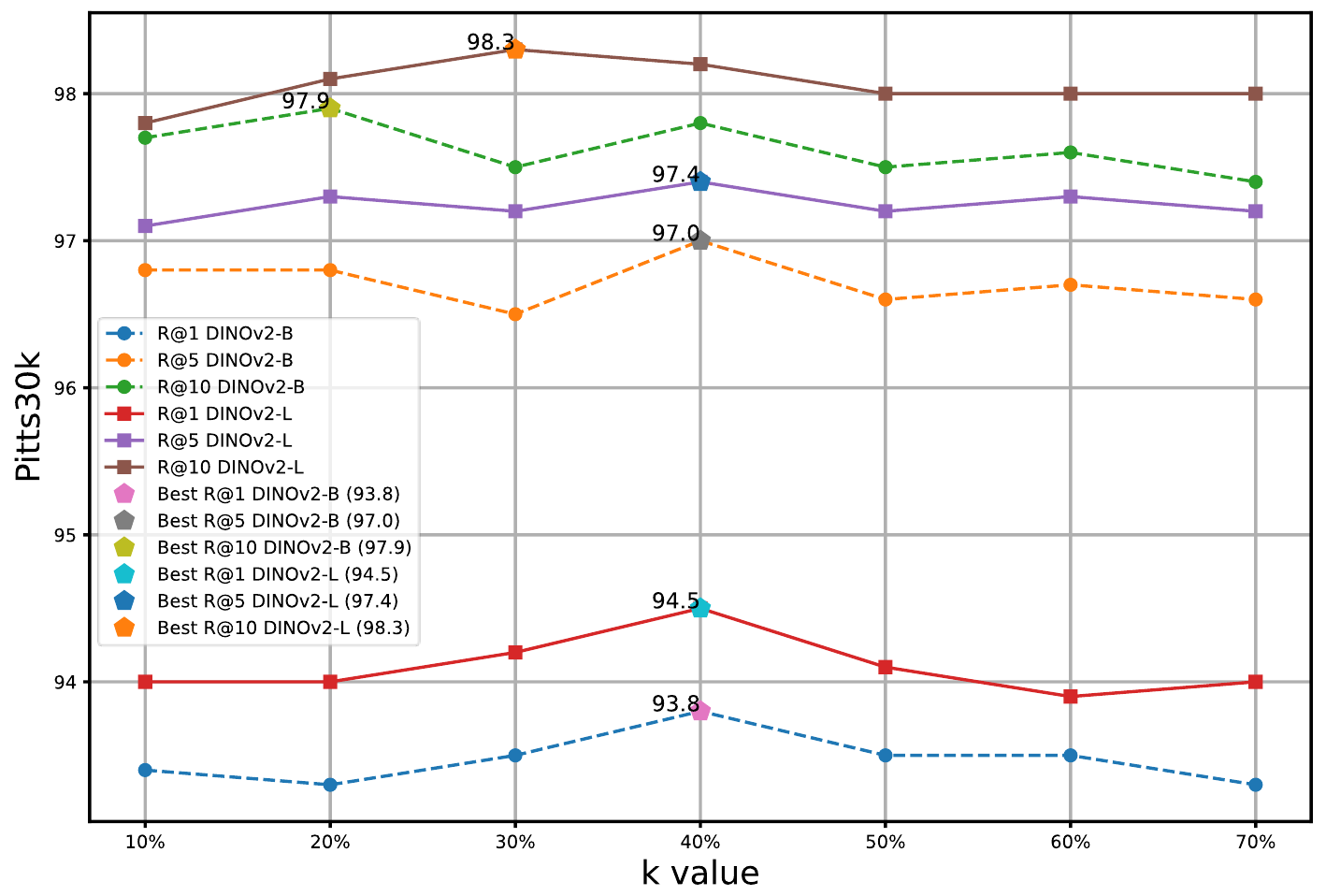}  
    \caption{{\bf Ablation study of the parameter $k$.} The best performance is achieved when $k$ is set to the top 40\% of the values in $\mathbf{M}$.}
    \label{fig:performance_plot}
    \vspace{-0.5cm}
\end{figure}

\begin{figure}[!h]
    \centering
    \includegraphics[width=\linewidth]{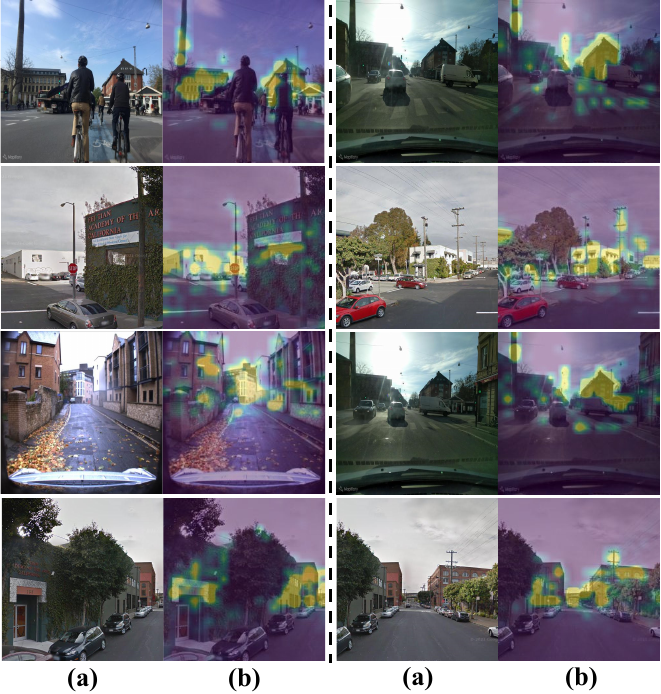}
\caption{Visualization of Discriminative Region. (a) shows the original image, and (b) shows the discriminative region mask $\mathbf{M}$. The heatmap emphasizes unique structures like buildings and vegetation, avoiding less informative areas such as the sky and roads.}
    \label{fig:mask_mix}
    \vspace{-0.5cm}
\end{figure}

\subsection{More Ablation Study}

In this section, we ablated $k$, which denotes the top positions in matrix $\mathbf{M}$ that are converted to 1 in the binary matrix $\mathbf{M}_{\text{bin}}$. As shown in the Figure \ref{fig:performance_plot}, we evaluated the impact of varying $k$ values on our method's performance, with $k$ ranging from 10\% to 70\% of the elements in $\mathbf{M}$.
Each line in the plot represents a different ranking performance metric, highlighting how the choice of $k$ influences the discriminative power of FoL. Notably, we annotated the best performance points on the plot to emphasize the optimal $k$ values.

\subsection{Visualization}

\noindent\textbf{Visualization of Discriminative Region.}
As shown in the Figure \ref{fig:mask_mix}, we focus on visualizing the importance of local features within various scenes. In part (a), the original image is presented, while part (b) illustrates the discriminative region mask \(\mathbf{M}\). The heatmap highlights critical elements within the scene, such as distinctive structures like buildings and vegetation, while deliberately minimizing attention to less informative regions, such as the sky and roads, which contribute little to the overall scene understanding.

\end{document}